\documentclass[pdflatex,sn-mathphys-num]{sn-jnl}% Math and Physical Sciences Numbered Reference Style
\usepackage{graphicx}%
\usepackage{multirow}%
\usepackage{amsmath,amssymb,amsfonts}%
\usepackage{amsthm}%
\usepackage{mathrsfs}%
\usepackage[title]{appendix}%
\usepackage{xcolor}%
\usepackage{textcomp}%
\usepackage{manyfoot}%
\usepackage{booktabs}%
\usepackage{algorithm}%
\usepackage{algorithmicx}%
\usepackage{algpseudocode}%
\usepackage{listings}%
\theoremstyle{thmstyleone}%
\theoremstyle{thmstyletwo}%

\theoremstyle{thmstylethree}%

\begin{document}

\title[Article Title]{Do large language models scrutinise what they review? A multimodal audit of scoring calibration, error detection, and author-identity effects}

%%=============================================================%%
%% GivenName	-> \fnm{Joergen W.}
%% Particle	-> \spfx{van der} -> surname prefix
%% FamilyName	-> \sur{Ploeg}
%% Suffix	-> \sfx{IV}
%% \author*[1,2]{\fnm{Joergen W.} \spfx{van der} \sur{Ploeg} 
%%  \sfx{IV}}\email{iauthor@gmail.com}
%%=============================================================%%

\author*[1]{\fnm{Emad} \sur{Alharbi}}\email{emalharbi@ut.edu.sa}

\affil*[1]{\orgdiv{Department of Information Technology}, \orgname{University of Tabuk}, \orgaddress{\city{Tabuk}, \postcode{71491}, \state{Tabuk}, \country{Saudi Arabia}}}

%%==================================%%
%% Sample for unstructured abstract %%
%%==================================%%

\abstract{Large language models (LLMs) are increasingly used to generate peer reviews, prompting examination of their capacity for critical evaluation. This study evaluates two multimodal LLMs, Qwen2.5-VL-72B and Pixtral-Large-124B, as reviewers across 165 submissions to the 2026 International Conference on Learning Representations, a venue that postdates both models' training cutoffs. Manuscripts were presented to both models with author identities blinded, replaced with high-prestige affiliations, or replaced with low-prestige affiliations, and in either text-only or text-with-figure format. Additionally, 145 verifiably detectable errors were inserted into 55 manuscripts to assess error identification under natural and verification-oriented prompts. Across all manuscript groups, including rejected submissions, LLM scores ranged from 7.0 to 8.1, whereas human mean scores ranged from 3.4 to 6.8. The models detected 12.1\% of the verified errors under natural prompting, and a one-sentence verification instruction increased detection to 22.2\%; however, 78\% of the errors remained undetected. Providing figures reduced error detection while increasing review scores. No visual error was reliably verified against its corresponding figure, and half of the text-only reviews described figures that were not provided. Author identity did not influence either review scores or error detection. LLM editorial decisions exactly matched those produced by simple score averaging.}


%%\pacs[JEL Classification]{D8, H51}

%%\pacs[MSC Classification]{35A01, 65L10, 65L12, 65L20, 65L70}

\maketitle

\section{Background}\label{sec1}
The 2026 International Conference on Learning Representations (ICLR 2026) received 19,814 submissions, requiring approximately 80,000 review reports, and reviewer recruitment has become a resource constraint across venues \cite{aczel2021billion, shah2022challenges}. Large language models (LLMs) have been used to accelerate the review process, either through reviewers informally using them to draft or refine reviews, or through venues piloting LLM-assisted reviewing \cite{liang2024monitoring,zhuang2025large}. Previous studies show that LLM-generated feedback overlaps substantially with human reviewer comments and is perceived as helpful by authors \cite{liang2024can, thakkar2026large}. However, the quality of LLM-generated reviews is a distinct question from whether authors perceive their comments as helpful. 

A separate concern is bias. Human peer review exhibits prestige effects \cite{peters1982peer,tomkins2017single,tomkins2017reviewer}, and earlier evaluations of LLM reviewers have identified similar affiliation biases \cite{von2024affiliation}. Existing evaluations of LLM reviewing are subject to four methodological weaknesses. First,\emph{contamination}: models are evaluated on venues whose papers may have been included in their training data, enabling retrieval rather than genuine evaluation. Second, \emph{version mismatch}: because OpenReview restricts revision histories to authors and organizers, the publicly available PDF of an accepted paper is the camera-ready version rather than the version reviewed by human reviewers. Consequently, studies based on accepted papers compare LLM output with human reviews of a different manuscript. To our knowledge, neither the prevalence nor the consequences of this mismatch have been quantified. Third, \emph{unverified error benchmarks}: error-injection studies rarely establish that injected errors are detectable in principle \cite{liu2023reviewergpt}. They often fail to preserve contradictory evidence in the manuscript or provide verified answer keys, making detection rates difficult to interpret. Fourth, \emph{modality blindness}: although peer review relies on figures, evaluations of LLM reviewers remain almost exclusively text-based. The performance of multimodal models on complete manuscripts, including whether they comment on unseen figures, has not been measured.

This paper addresses all four methodological weaknesses. Our corpus consists of ICLR 2026 submissions, which postdate the training cutoffs of both reviewer models. To our knowledge, this study is the first version-provenance census of an OpenReview venue, showing that pre-review versions of accepted manuscripts are almost entirely unavailable from public sources. The original versions were recovered and authenticated using line-anchored quotations from their corresponding reviews, and all human-comparison analyses were restricted to version-matched manuscripts. A 145-error benchmark was constructed in which each injected error was verified to be detectable. Contradictory evidence was retained after each edit and labelled with an empirically verified subtype, including a certified visual subset answerable only from figures. Author identity (blind, high-prestige, or low-prestige), input modality (text only or text with figures), manuscript integrity (clean or error-injected), and review prompt (natural or verification-oriented) were manipulated in a within-manuscript factorial design across 165 stratified manuscripts, generating 9,900 structured reviews from two frontier open-weight multimodal LLMs. The resulting reviews were adjudicated by an LLM editor and scored against a frozen answer key by an independent LLM judge, with stratified human validation.

This experimental design examines whether LLM scores align with human assessments of manuscript quality, whether author identity affects review scores or error detection, whether access to figures improves review quality, and whether an LLM editor provides value beyond simple score aggregation. The benchmark, corpus provenance labels, and evaluation pipeline are released to support reproducibility \url{https://doi.org/10.5281/zenodo.21720411}. 
\section{Methods}\label{sec2}
\subsection{Study design}
Peer review was conducted using LLMs as reviewers. Each manuscript was reviewed under two input conditions: text only and text with figures. Each input condition was further evaluated using blinded author identities and high- and low-prestige affiliations. In addition, errors were introduced into the manuscript to evaluate review integrity. Two review prompts, a natural prompt and an error-detection prompt, were used alongside two LLMs: one served as reviewer and editor, and the other acted as a judge for scoring error detection.

\subsection{Data collection}
A total of 19,814 ICLR 2026 submissions were collected via the OpenReview API, along with their official reviews and submission decisions; 14,175 had received a final decision. ICLR 2026 was selected because its September 2025 submission deadline postdates the training data cut-off for both LLMs, ensuring that its submissions were unavailable during LLM training. All datasets are publicly available.

\subsection{Data version and recovery}
For ICLR 2026, OpenReview restricts submission revision histories to authors and organizers. Consequently, the publicly available PDF of an accepted manuscript is the camera-ready version rather than the version reviewed by human reviewers, whereas the publicly available version of a rejected manuscript is the human-reviewed version. To recover the reviewed versions of accepted manuscripts, the Internet Archive was queried for snapshots captured during the review period, yielding 16 manuscripts.

\subsection{Data sampling and anonymization \label{sec:anonymisation}}
A seed-frozen stratified random sample of 150 manuscripts was selected from decided submissions with at least three official reviews, parseable scores, a usable PDF, and no more than 40 pages: 50 clear accepts with a mean review score of at least 6.5, 50 borderline manuscripts with a mean review score between 5.0 and 6.5, and 50 clear rejects with a mean review score below 4.5. An additional 15 manuscripts retrieved from the Internet Archive, one of which had already been selected as a clear accept, increased the final sample to 165 for the experiment. 

The manuscripts were anonymized by removing author blocks, acknowledgments, and reference lists. Acknowledgment and reference list sections were removed to prevent author identities from being inferred through institutional acknowledgments or self-citations. Page headers were cropped to remove the publication venue from camera-ready versions.
\subsection{Experimental conditions}
18 experiments were conducted, as shown in Table~\ref{tab:experiments}, with the factors described below. 
\subsubsection{Author identity}
Authors' identities were removed from manuscripts for blind review, along with the acknowledgment and reference list sections, as described in Section~\ref {sec:anonymisation}. For the high-prestige review bias test, authors were assigned to highly ranked institutions and corresponding computer science departments. For the low-prestige review bias test, authors were assigned fictitious institutional affiliations with names resembling small institutions.
\subsubsection{Manuscript integrity}
The manuscripts were injected with three types of errors using find-and-replace edits: A) numeric contradictions, in which a value in the abstract or introduction was changed to conflict with the results section or a table; B) logical or trend contradictions, in which conclusions, dataset properties, or effect directions were reversed while other statements remained unchanged; and C) figure--text mismatches, in which figure descriptions were rewritten to contradict the corresponding figures. A subset of manuscripts contained errors that appeared nowhere else in the text and could only be detected from the figures. Table \ref{tab:subtypes} presents the subtypes within these error categories, along with examples of original and injected errors. The errors were proposed by an LLM (Pixtral-Large-Instruct-2411), then implemented and verified through automated pre-screening. The errors were reviewed to ensure that they were detectable by reviewers.

\subsubsection{Reviewing prompt design}
Two prompts were used: a natural prompt and a verification-oriented prompt that included explicit cross-checking instructions. The verification-oriented prompt explicitly emphasized checking abstract numbers against results and tables, comparing figure claims with images, and confirming that figure and table references (e.g. `as shown in Figure 3') corresponded to content consistent with their description.

\subsubsection{Review generation and evaluation pipeline}
Qwen2.5-VL-72B-Instruct and Pixtral-Large-Instruct-2411 were used to generate reviews, with inference performed on 8×A100 GPUs. The reviews were stored in JSON format and included an overall score (1--10), itemized strengths and weaknesses with locations, and a field for a figure or table critique, which was instructed to state `not assessable' when figures were unavailable. Each review was generated three times to reduce random bias, and the final score was calculated as the mean of the three runs using a temperature of 0.3 and a top-p value of 0.9.

An LLM editor (Qwen2.5-VL-72B) adjudicated each pair of natural-prompt reviews to produce an accept-or-reject decision, which was compared with the mean reviewer score and the venue decision. An LLM judge (Pixtral-Large) evaluated every review of the error-injected manuscript against each injected error, considering an error detected only if the review correctly identified both its location and nature, supported by quoted evidence. The judge model belonged to a different model family from one reviewer and applied an identical rubric to both reviewers' outputs without knowledge of their authorship. A sample of 63 judgments, balanced across reviewer models, prompt variants, and verdicts, was independently re-evaluated by an author, yielding an agreement of 92.1\% with comparable agreement for both reviewers' outputs (95.8\% and 89.7\% for Qwen2.5-VL-72B and Pixtral-Large, respectively).

\subsubsection{Statistical analysis}
The results were analysed using binomial generalized estimating equations (GEE) \cite{liang1986longitudinal}, clustered by errors (145 clusters) to estimate the effect of each factor on the odds of error detection. In addition, each manuscript's mean LLM score was correlated with its mean human review score.

\begin{table}[t]
\centering
\caption{Experiments performed and their comparisons. Each experiment was conducted across all applicable manuscripts using both reviewer models, with three replications. Each group was compared with an otherwise identical group differing only in the factor under investigation.}
\label{tab:experiments}
\footnotesize
\setlength{\tabcolsep}{4pt}
\begin{tabular}{l l l l l l l}
\toprule
ID & Version & Modality & Identity & Prompt & Compares & Outcome \\
\midrule
E1  & Clean & text     & blind & natural & human reviews & \multirow{2}{*}{\shortstack[l]{Score agreement with human\\ reviews}} \\
E2  & Clean & text+fig & blind & natural & human reviews & \\
\addlinespace
E3  & Clean & text     & high-prestige & natural & E1 & \multirow{2}{*}{High-prestige score bias.} \\
E4  & Clean & text+fig & high-prestige & natural & E2 & \\
\addlinespace
E5  & Clean & text     & low-prestige & natural & E1 & \multirow{2}{*}{Low-prestige score bias.} \\
E6  & Clean & text+fig & low-prestige & natural & E2 & \\
\midrule
E7  & Error-inj. & text     & blind & natural & answer key & \multirow{6}{*}{\shortstack[l]{Detection rate by error subtype\\ and identity effect on \\ detection (E9--E12).}} \\
E8  & Error-inj. & text+fig & blind & natural & answer key & \\
E9  & Error-inj. & text     & high-prestige & natural & E7 & \\
E10 & Error-inj. & text+fig & high-prestige & natural & E8 & \\
E11 & Error-inj. & text     & low-prestige & natural & E7 & \\
E12 & Error-inj. & text+fig & low-prestige & natural & E8 & \\
\midrule
E13 & Error-inj. & text     & blind & verification & E7  & \multirow{6}{*}{\shortstack[l]{Verification-prompt effect\\ on detection.}} \\
E14 & Error-inj. & text+fig & blind & verification & E8  & \\
E15 & Error-inj. & text     & high-prestige & verification & E9  & \\
E16 & Error-inj. & text+fig & high-prestige & verification & E10 & \\
E17 & Error-inj. & text     & low-prestige & verification & E11 & \\
E18 & Error-inj. & text+fig & low-prestige & verification & E12 & \\
\bottomrule
\end{tabular}
\end{table}

\begin{table}[t]
\centering
\caption{Error categories and their subtypes. For each error subtype, an example of the original text and the corresponding injected errors is shown.}
\label{tab:subtypes}
\footnotesize
\setlength{\tabcolsep}{4pt}
\begin{tabular}{l l p{3.4cm} p{3.4cm} p{3.6cm}}
\toprule
Category & Subtype & Definition & Original & Injected \\
\midrule
\multirow{3}{*}{\shortstack[l]{A}}
 & Numeric & Value in the abstract or introduction altered to contradict the same quantity in the results or a table.
 & `a 2.25$\times$ speed-up' (results also report 2.25$\times$)
 & `a 2.5$\times$ speed-up' \\
\addlinespace
 & Pvalue & Statistical claim rendered internally invalid.
 & `significantly outperforms the baseline'
 & `significantly outperforms the baseline (p = 0.08)' \\
\addlinespace
 & Claim & Qualitative claim strengthened beyond what the results support.
 & `performs comparably to prior methods'
 & `consistently surpasses all prior methods' \\
\midrule
\multirow{2}{*}{\shortstack[l]{B}}
 & Logical & Stated conclusion, dataset property, or methodological statement reversed while corroborating statements elsewhere remain intact.
 & `we train on distinct splits with no overlap' (the setup section describes the distinct splits)
 & `we train and evaluate on the same split' \\
\addlinespace
 & Trend & Stated direction of effect reversed.
 & `accuracy improves with depth' (tables show the improvement)
 & `accuracy degrades with depth' \\
\midrule
\multirow{3}{*}{\shortstack[l]{C}}
 & Figure contradiction & Sentence describing a figure rewritten to contradict the figure's content.
 & `Figure 4 shows the loss decreasing monotonically'
 & `Figure 4 shows the loss increasing after epoch 10' \\
\addlinespace
& Text corroborated & As C\_figure\_contradiction, with the contradiction additionally corroborated by other text (e.g. a table).
 & `Figure 2 shows accuracy rising with model size' (Table 3 reports the rising accuracy values)
 & `Figure 2 shows accuracy falling beyond 7B parameters' \\
\addlinespace
 & Pure visual & Altered visual claim appears nowhere else in the text; only the figure image contradicts it (certified subset, $n=9$).
 & `the distribution in Figure 5 is symmetric' (no other sentence describes the shape)
 & `the distribution in Figure 5 is strongly right-skewed' \\
\bottomrule
\end{tabular}
\end{table}
\section{Results}
\subsection{Score calibration against human reviewers}
\begin{figure}[h]
\centering
\includegraphics[width=0.9\textwidth]{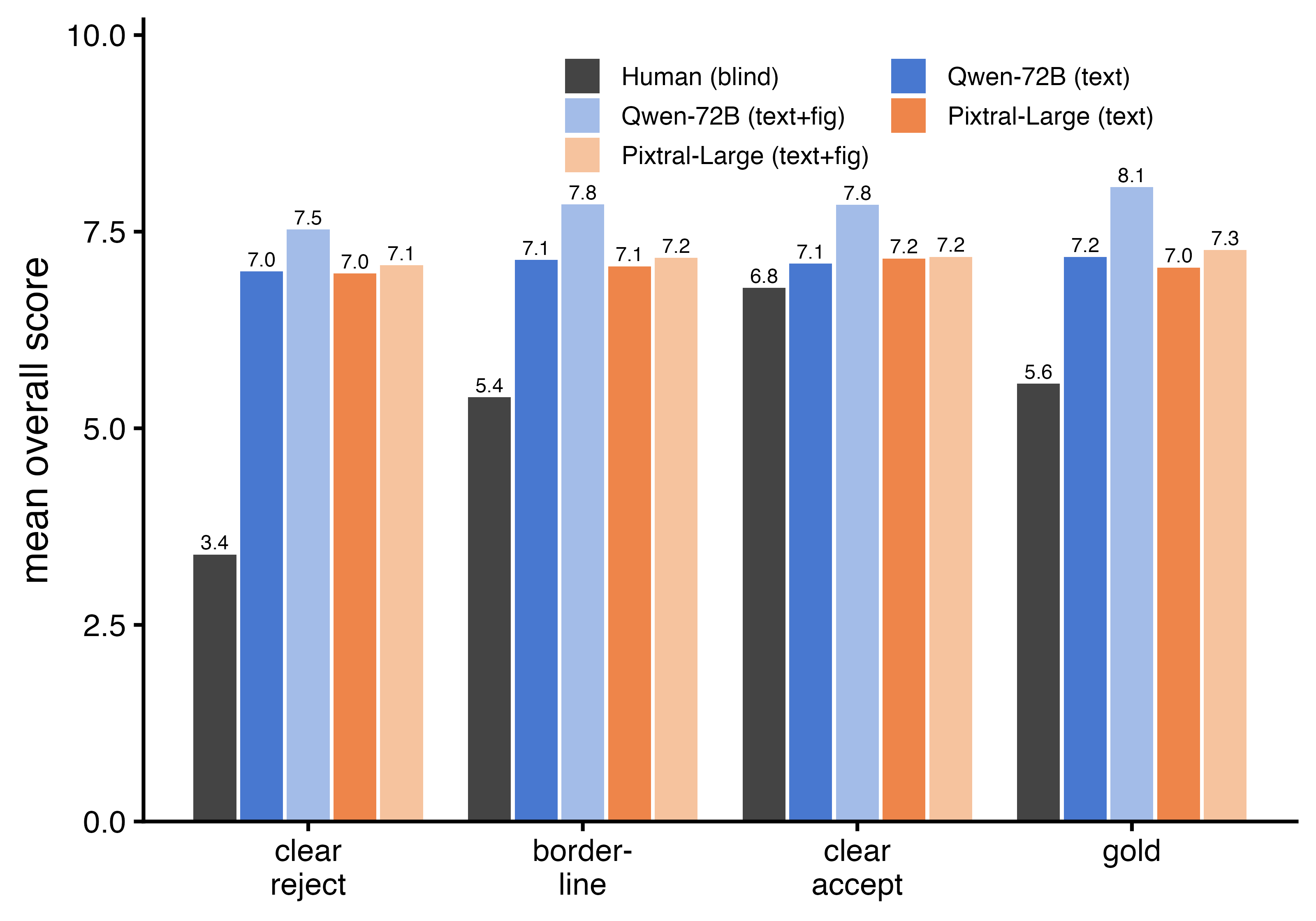}
\caption{Mean review score for the two LLMs and human reviewers for the four groups of manuscripts}\label{fig:Mean_review_score}
\end{figure}

Figure \ref{fig:Mean_review_score} shows the mean review scores assigned by the two LLMs and human reviewers under blinded author identity, separately for text-only and text-with-figure inputs. Human reviewers assigned mean scores of 3.4 to clear rejections, 5.4 to borderline manuscripts, and 6.8 to clear accepts. In contrast, the mean text-only scores assigned by both LLMs under blinded conditions ranged narrowly from 7.0 to 7.2 across all manuscript groups. With figures provided, Qwen2.5-VL assigned scores ranging from 7.5 to 8.1, whereas Pixtral-Large assigned scores between 7.1 and 7.3. Providing figures increased scores by 0.5 to 0.9 points for Qwen2.5-VL and by 0.1 to 0.2 points for Pixtral-Large across all manuscript groups. The gold manuscripts, which received a mean human score of 5.6, achieved the highest Qwen2.5-VL score of any manuscript group, reaching 8.1 for text with figures.
\subsection{Manuscript error detection}
\begin{figure}[h]
\centering
\includegraphics[width=0.9\textwidth]{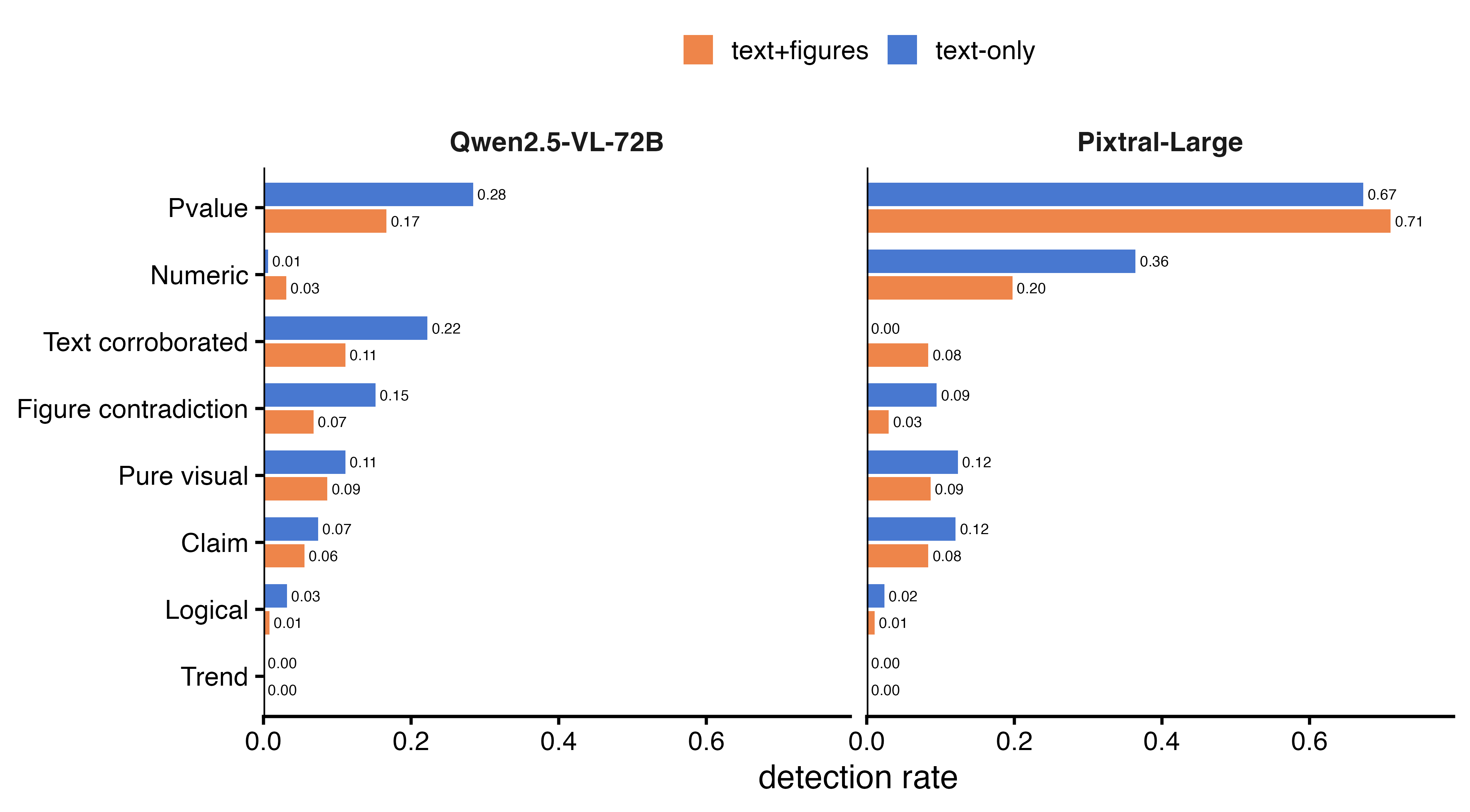}
\caption{Fraction of error detection by error subtypes for Qwen2.5-VL and Pixtral-Large models}\label{fig:detection_by_subtype}
\end{figure}

Figure \ref{fig:detection_by_subtype} shows the detection rates for the 145 verified errors under the natural prompt, stratified by the error subtypes listed in Table \ref{tab:subtypes}, for the two review models. Invalid statistical claims and p-value errors were detected most frequently, with detection rates of 0.67 (text only) and 0.71 (text with figures) for Pixtral-Large, compared with 0.28 and 0.17 for Qwen2.5-VL. In contrast, Qwen2.5-VL detected almost no numeric contradiction errors, with rates of 0.01 for text only and 0.03 for text with figures, whereas Pixtral-Large achieved its second-highest detection rates of 0.36 and 0.20 for text only and text with figures, respectively. Trend-reversal errors were not detected by either model under either input condition. Error detection was significantly lower when figures were provided than under the text-only condition (odds ratio = 1.53, ($ p=7.9\times10^ {-7}$), with GEE clustered by error).

\subsection{Effect of verification-oriented prompting}
\begin{figure}[h]
\centering
\includegraphics[width=0.9\textwidth]{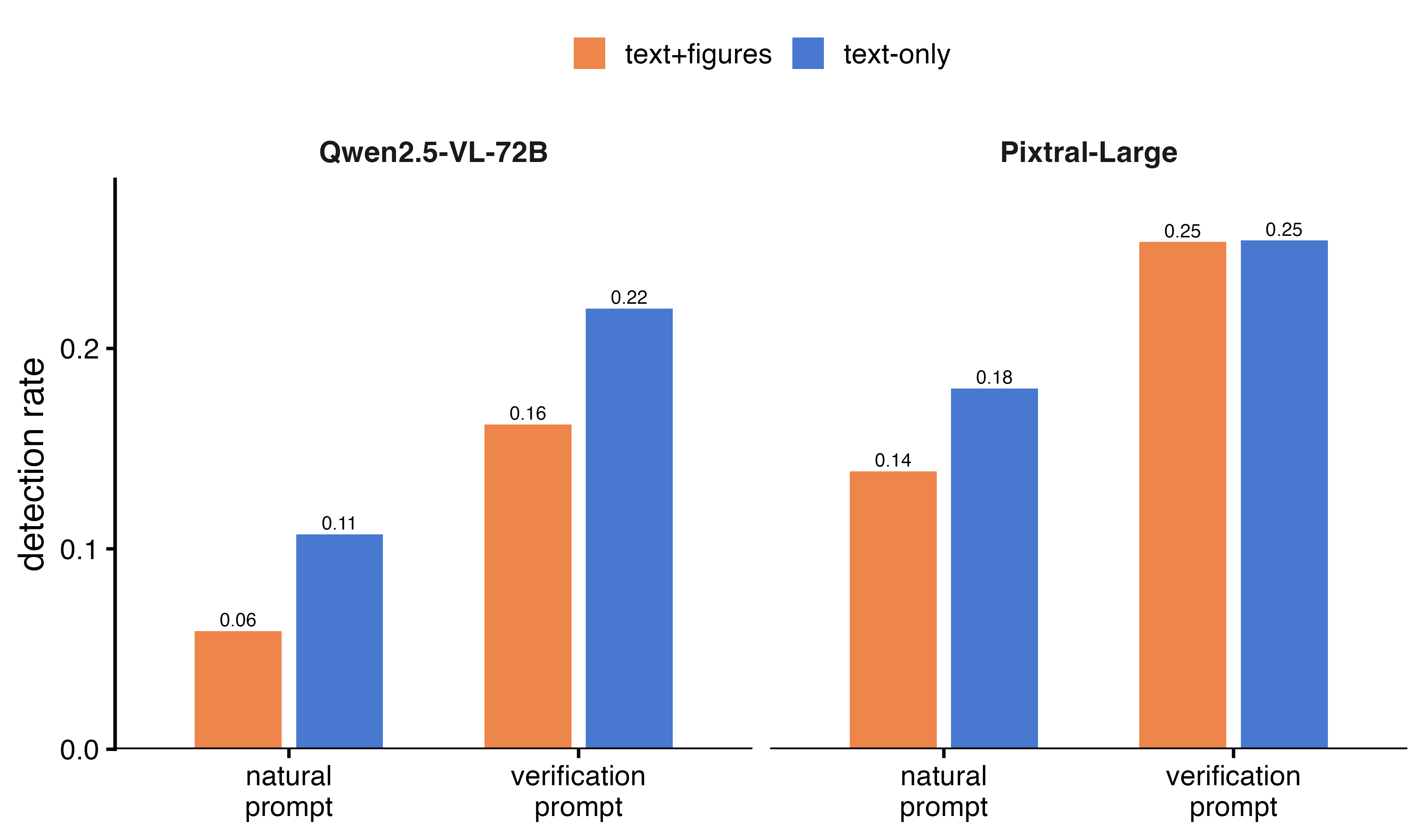}
\caption{Fraction of error detection using two versions of the prompt: natural and verification prompts for the Qwen2.5-VL and Pixtral-Large models}\label{fig:prompt_effect}
\end{figure}

Figure \ref{fig:prompt_effect} shows the error detection rates for the two LLMs under the natural and verification-oriented prompts. The verification-oriented prompt increased the overall error detection rate across both LLMs from 12.1\% to 22.2\% (odds ratio = 2.05, $p = 1.3\times10^{-12}$); the Qwen2.5-VL detection increased from 0.06 to 0.16 for text with figures and from 0.11 to 0.22 for text only. For Pixtral-Large, the corresponding rates increased from 0.14 to 0.25 and from 0.18 to 0.25, respectively. Under the verification-oriented prompt, Pixtral-Large achieved the same error detection rate (0.25) for text-only and text-with-figure inputs.

\subsection{Effects of author identity on scores and error detection}
\begin{figure}[h]
\centering
\includegraphics[width=0.9\textwidth]{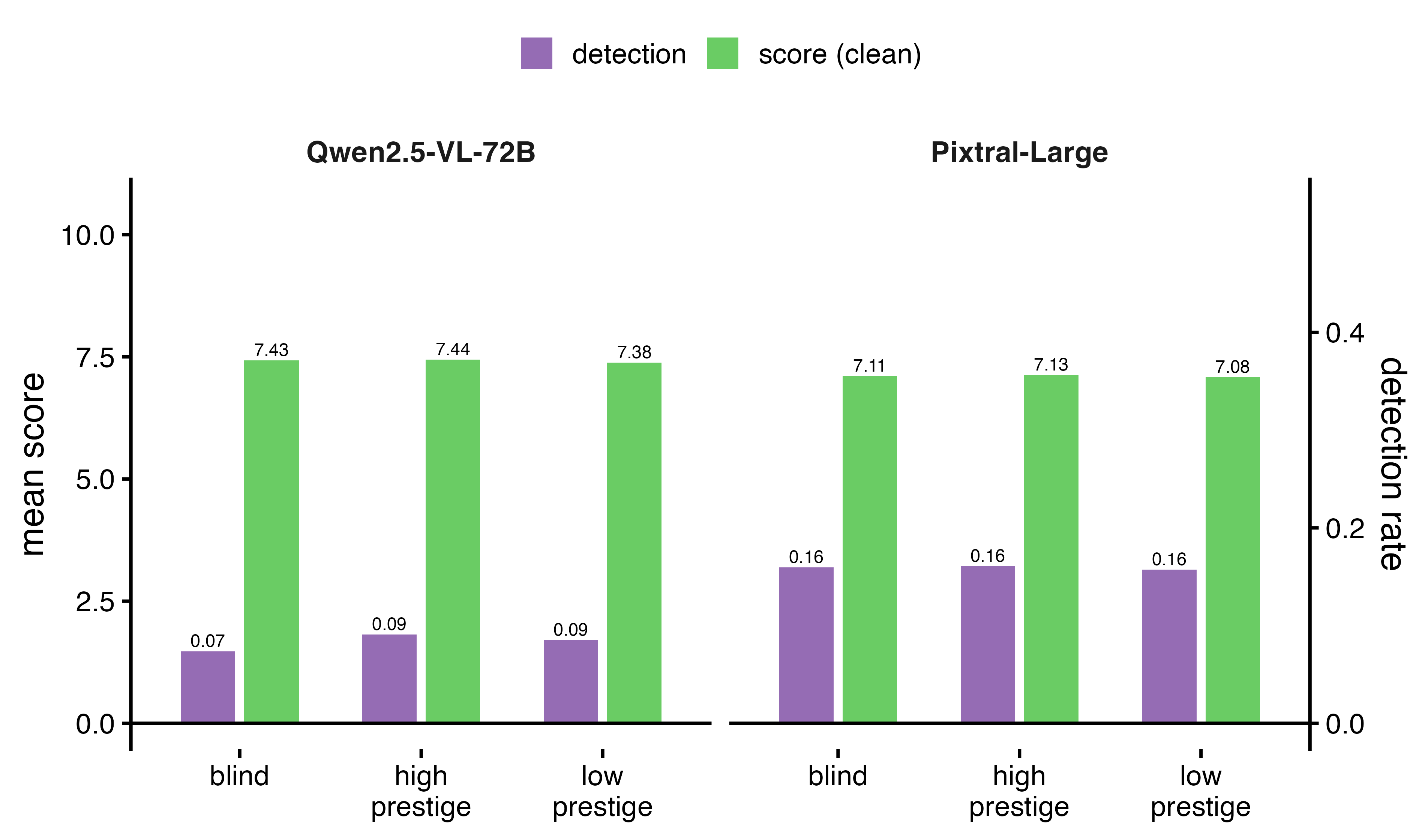}
\caption{Author identity effect on the error detection and review mean score.}\label{fig:figD_identity_nulls}
\end{figure}

Figure \ref{fig:figD_identity_nulls} shows the mean review scores and error detection rates for the LLMs across author identity conditions. Neither outcome differed across identity conditions for either model. Relative to blinded review, the GEE yielded odds ratios of 1.09 (p = 0.19) for high-prestige identities and 1.04 (p = 0.49) for low-prestige identities. The mean Qwen2.5-VL scores were 7.43 for blinded author identities, 7.44 for high-prestige author identities, and 7.38 for low-prestige author identities; the corresponding Pixtral-Large scores were 7.11, 7.13, and 7.08, respectively. Therefore, author affiliation affected neither the scores assigned to clean manuscripts nor the detection of errors in flawed manuscripts by either model.

\subsection{Editorial decisions}
The LLM editor, Qwen2.5-VL-72B, received each pair of natural-prompt reviews, along with the full manuscript, and produced an accept-or-reject decision accompanied by a written assessment. The editor's decisions aligned with the venue's decisions for 60.6\% of the review pairs. This agreement was identical to that of a naive baseline, which accepted any manuscript with a mean reviewer score of at least 6.
\section{Discussion}
\subsection{Principal findings}
The study generated nearly 10,000 reviews, evaluated two multimodal LLMs, and assessed performance using 145 verified injected errors. The two LLMs assigned consistently higher scores than human reviewers, and their mean scores showed only weak correlations with human judgment for Qwen2.5-VL and Pixtral-Large. Under natural prompting, both LLMs detected only a small fraction of the injected errors, and providing figures reduced rather than improved error detection. A one-sentence verification instruction approximately doubled error detection, whereas author identity had minimal effect on review scores. These findings indicate limitations in LLM performance rather than model bias. Author identity did not influence model behaviour, and no evidence of affiliation bias was observed in either review scores or scrutiny.
\subsection{Relation to prior work}
Studies of LLM-generated review feedback have reported substantial overlap with human comments and high perceived utility \cite{liang2024can,thakkar2026large}. Although LLMs can generate useful review comments, their scoring is of limited value, and their error-detection capabilities may be inadequate. Comparison with human reviewers showed that LLM reviews identified only surface-level weaknesses and failed to detect conceptual deficiencies. 
\subsection{Implications}
Current LLMs may be useful as comment generators under human oversight but are not suitable as scorers, screeners, or figure checkers. Verification-oriented prompting is inexpensive and should be the default in assisted-review pilots, although it is not a complete solution. Ensemble review using heterogeneous models may outperform either model alone, even when one model performs poorly on numeric cross-checking. For the meta-research community, the released benchmark of subtype-labelled, detectability-verified errors, along with the provenance-audited corpus, provides reusable resources applicable to any venue with publicly available reviews.
\subsection{Limitations}
The evaluation included two LLMs from different model families and manuscripts from a single venue, research field, and review format. One LLM served as the judge for error detection, and one reviewer belonged to the same model family. This limitation was mitigated through symmetric, authorship-blind evaluation and human validation. Additionally, the low-prestige institutional names were fictitious, thereby testing unfamiliarity combined with modest naming rather than affiliation with an established low-prestige institution.

\section{Conclusions}
This study assessed the reliability of multimodal LLMs in peer review by evaluating their performance across controlled review conditions and verified error-detection tasks. The results indicate that LLM reviewers failed to detect errors, assigned inflated review scores, and described figures that were not provided. Author identity had minimal effect on the LLM reviews, whereas providing figures alongside manuscript text slightly increased review scores while reducing error detection. These findings suggest that until LLMs demonstrate reliable capacity for critical scrutiny, they should serve as complements to human reviewers rather than replace them.
%%===================================================%%
%% For presentation purpose, we have included        %%
%% \bigskip command. Please ignore this.             %%
%%===================================================%%

%%=============================================%%
%% For submissions to Nature Portfolio Journals %%
%% please use the heading ``Extended Data''.   %%
%%=============================================%%

%%=============================================================%%
%% Sample for another appendix section			       %%
%%=============================================================%%

%% \section{Example of another appendix section}\label{secA2}%
%% Appendices may be used for helpful, supporting or essential material that would otherwise 
%% clutter, break up or be distracting to the text. Appendices can consist of sections, figures, 
%% tables and equations etc.

%%===========================================================================================%%
%% If you are submitting to one of the Nature Portfolio journals, using the eJP submission   %%
%% system, please include the references within the manuscript file itself. You may do this  %%
%% by copying the reference list from your .bbl file, paste it into the main manuscript .tex %%
%% file, and delete the associated \verb+\bibliography+ commands.                            %%
%%===========================================================================================%%

\bibliography{sn-bibliography}% common bib file

@article{liang2024can,
  title={Can large language models provide useful feedback on research papers? A large-scale empirical analysis},
  author={Liang, Weixin and Zhang, Yuhui and Cao, Hancheng and Wang, Binglu and Ding, Daisy Yi and Yang, Xinyu and Vodrahalli, Kailas and He, Siyu and Smith, Daniel Scott and Yin, Yian and others},
  journal={NEJM AI},
  volume={1},
  number={8},
  pages={AIoa2400196},
  year={2024},
  publisher={Massachusetts Medical Society}
}

@article{aczel2021billion,
  title={A billion-dollar donation: estimating the cost of researchers’ time spent on peer review},
  author={Aczel, Balazs and Szaszi, Barnabas and Holcombe, Alex O},
  journal={Research integrity and peer review},
  volume={6},
  number={1},
  pages={14},
  year={2021},
  publisher={Springer}
}

@article{peters1982peer, title={Peer-review practices of psychological journals: The fate of published articles, submitted again}, volume={5}, DOI={10.1017/S0140525X00011183}, number={2}, journal={Behavioral and Brain Sciences}, author={Peters, Douglas P. and Ceci, Stephen J.}, year={1982}, pages={187–195}}

@article{von2024affiliation,
  title={Affiliation bias in peer review of abstracts by a large language model},
  author={von Wedel, Dario and Schmitt, Rico A and Thiele, Moritz and Leuner, Raphael and Shay, Denys and Redaelli, Simone and Schaefer, Maximilian S},
  journal={JAMA},
  volume={331},
  number={3},
  pages={252--253},
  year={2024}
}

@article{shah2022challenges,
  title={Challenges, experiments, and computational solutions in peer review},
  author={Shah, Nihar B},
  journal={Communications of the ACM},
  volume={65},
  number={6},
  pages={76--87},
  year={2022},
  publisher={ACM New York, NY, USA}
}

@article{liang2024monitoring,
  title={Monitoring ai-modified content at scale: A case study on the impact of chatgpt on ai conference peer reviews},
  author={Liang, Weixin and Izzo, Zachary and Zhang, Yaohui and Lepp, Haley and Cao, Hancheng and Zhao, Xuandong and Chen, Lingjiao and Ye, Haotian and Liu, Sheng and Huang, Zhi and others},
  journal={arXiv preprint arXiv:2403.07183},
  year={2024}
}

@article{thakkar2026large,
  title={A large-scale randomized study of large language model feedback in peer review},
  author={Thakkar, Nitya and Yuksekgonul, Mert and Silberg, Jake and Garg, Animesh and Peng, Nanyun and Sha, Fei and Yu, Rose and Vondrick, Carl and Zou, James},
  journal={Nature Machine Intelligence},
  volume={8},
  number={3},
  pages={326--336},
  year={2026},
  publisher={Nature Publishing Group UK London}
}

@article{tomkins2017single,
  title={Reviewer bias in single-versus double-blind peer review},
  author={Tomkins, Andrew and Zhang, Min and Heavlin, William D},
  journal={Proceedings of the National Academy of Sciences},
  volume={114},
  number={48},
  pages={12708--12713},
  year={2017},
  publisher={National Academy of Sciences}
}

@article{liang1986longitudinal,
  title={Longitudinal data analysis using generalized linear models},
  author={Liang, Kung-Yee and Zeger, Scott L and others},
  journal={biometrika},
  volume={73},
  number={1},
  pages={13--22},
  year={1986}
}

@article{tomkins2017reviewer,
  title={Reviewer bias in single-versus double-blind peer review},
  author={Tomkins, Andrew and Zhang, Min and Heavlin, William D},
  journal={Proceedings of the National Academy of Sciences},
  volume={114},
  number={48},
  pages={12708--12713},
  year={2017},
  publisher={National Academy of Sciences}
}

@article{liu2023reviewergpt,
  title={Reviewergpt? an exploratory study on using large language models for paper reviewing},
  author={Liu, Ryan and Shah, Nihar B},
  journal={arXiv preprint arXiv:2306.00622},
  year={2023}
}

@article{zhuang2025large,
  title={Large language models for automated scholarly paper review: A survey},
  author={Zhuang, Zhenzhen and Chen, Jiandong and Xu, Hongfeng and Jiang, Yuwen and Lin, Jialiang},
  journal={Information Fusion},
  volume={124},
  pages={103332},
  year={2025},
  publisher={Elsevier}
}
%% if required, the content of .bbl file can be included here once bbl is generated
%%\input sn-article.bbl

\end{document}